\documentclass[10pt,twocolumn,letterpaper]{article}

\usepackage{multirow}

\usepackage[pagenumbers]{cvpr} % To force page numbers, e.g. for an arXiv version

\usepackage[dvipsnames]{xcolor}

\definecolor{cvprblue}{rgb}{0.21,0.49,0.74}
\usepackage[pagebackref,breaklinks,colorlinks,citecolor=cvprblue]{hyperref}

\def\paperID{*****} % *** Enter the Paper ID here
\def\confName{CVPR}
\def\confYear{2024}

\title{ExCeL : Combined Extreme and Collective Logit Information for Enhancing Out-of-Distribution Detection}

\author{Naveen Karunanayake\\
The University of Sydney\\
Australia\\
{\tt\small naveen.karunanayake@sydney.edu.au}
\and
Suranga Seneviratne\\
The University of Sydney\\
Australia\\
{\tt\small suranga.seneviratne@sydney.edu.au}
\and
Sanjay Chawla\\
Qatar Computing Research Institute\\
Qatar\\
{\tt\small schawla@hbku.edu.qa}
}

\begin{document}
\maketitle
\begin{abstract}
Deep learning models often exhibit overconfidence in predicting out-of-distribution (OOD) data, underscoring the crucial role of OOD detection in ensuring reliability in predictions. Among various OOD detection approaches, post-hoc detectors have gained significant popularity, primarily due to their ease of use and implementation. However, the effectiveness of most post-hoc OOD detectors has been constrained as they rely solely either on extreme information, such as the maximum logit, or on the collective information (i.e., information spanned across classes or training samples) embedded within the output layer. In this paper, we propose ExCeL that combines both extreme and collective information within the output layer for enhanced accuracy in OOD detection. We leverage the logit of the top predicted class as the extreme information (i.e., the maximum logit), while the collective information is derived in a novel approach that involves assessing the likelihood of other classes appearing in subsequent ranks across various training samples. Our idea is motivated by the observation that, for in-distribution (ID) data, the ranking of classes beyond the predicted class is more deterministic compared to that in OOD data. Experiments conducted on CIFAR100 and ImageNet-200 datasets demonstrate that ExCeL consistently is among the five top-performing methods out of twenty-one existing post-hoc baselines when the joint performance on near-OOD and far-OOD is considered (i.e., in terms of AUROC and FPR95). Furthermore, ExCeL shows the best overall performance across both datasets, unlike other baselines that work best on one dataset but has a performance drop in the other.

\end{abstract}    
\section{Introduction}
\label{sec:introduction}

\begin{figure}[t!]
    \centering
    \includegraphics[width=0.49\textwidth]{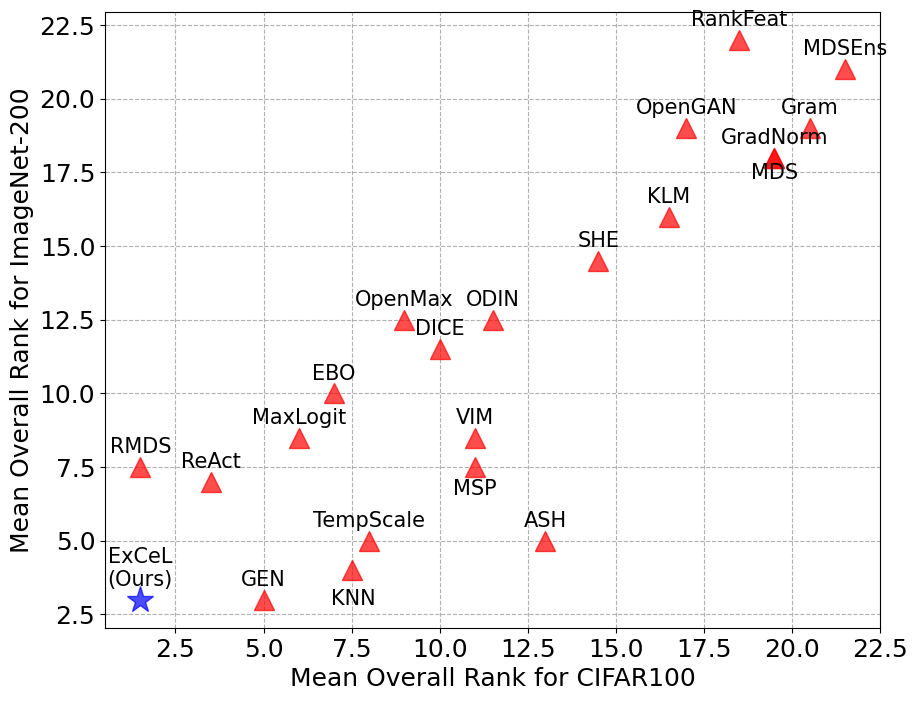}
    \caption{The performance comparison of 22 post-hoc OOD detection algorithms based on \textit{mean overall rank}. The x-axis represents the mean overall rank when CIFAR100 serves as ID, while y-axis represents the mean overall rank when ImageNet-200 is considered as ID. \textbf{ExCeL} shares the equal best performance on CIFAR100 and ImageNet-200 datasets with RMDS and GEN methods, respectively.}
    \label{fig:AUROC_figure}
\end{figure}

Deep neural networks (DNNs) deployed in the open world often encounter a diverse range of inputs from unknown classes, commonly referred to as out-of-distribution (OOD) data. However, DNNs' tendency to be overly confident, yet inaccurate about such inputs makes them less reliable, particularly in safety-critical applications such as autonomous driving~\cite{filos2020can} and healthcare~\cite{roy2022does}. Therefore, a DNN should be able to identify and avoid making predictions on these OOD inputs that differ from its training data. For instance, in an autonomous vehicle, the driving system must promptly alert the driver and transfer control when it detects unfamiliar scenes or objects that were not encountered during its training~\cite{nitsch2021out}. Accordingly, addressing the challenge of OOD detection has gained significant attention in recent studies~\cite{yang2021generalized}.

Among various OOD detection methods, post-hoc inference techniques~\cite{openmax,msp,tempscale,odin,mds,rmds,gram,ebo,opengan,gradnorm,react,mls,vim,knn,dice,rankfeat,ash,she} stand out as they can be applied to any pre-trained model, making them versatile and applicable to a wide range of models without the need for modifications during the training phase. These techniques extract crucial information from intermediate~\cite{gram, nmd, gradnorm} or output~\cite{msp, mls, ebo, mds, rmds} layers of DNNs, establishing an OOD score to distinguish between in-distribution (ID) and OOD samples. Since the output layer (i.e., normalised or unnormalised probabilities) in a DNN adeptly captures high-level semantics (i.e., objects, scenes etc.), researchers have increasingly focused on employing its features for OOD detection. For example, Hendrycks et al.~\cite{msp} proposed the \textit{maximum} softmax probability (MSP) as an initial baseline for OOD detection. Subsequently, Hendrycks et al.~\cite{mls} and Liu et al.~\cite{ebo} further harnessed more \textit{extreme} information from the output layer for OOD detection, utilising the \textit{maximum} logit and energy (i.e., a smooth approximation for the maximum logit), respectively. Moreover, recent studies have incorporated advanced post-processing techniques~\cite{odin, dice, react} to enhance the performance exhibited by MSP and energy scores. While all these methods focus on extreme information, some studies have adopted a broader perspective, considering information spanned across ID classes or training samples~\cite{mls, knn, mds, rmds}, which we refer to here as \textit{collective} information. For example, Lee et al.~\cite{mds} proposed a method to fit a class-conditional Gaussian distribution on the penultimate layer features using the \textit{training samples} and derive an OOD score with Mahalanobis distance.

In this paper, we introduce a novel metric ExCeL, designed to enhance OOD detection by incorporating both extreme and collective information at the output layer. While the logit of the top predicted class, commonly referred to as \textit{max logit}~\cite{mls} captures the extreme information, we show that the likelihood of other classes in subsequent ranks yields the collective information required to improve the distinguishability of OOD samples. Our approach is motivated by the observation that, during inference, when an input is predicted as a specific ID class, the rankings of the remaining classes are more consistently predictable for ID data compared to OOD data. Therefore, each ID class can be characterised by a unique \textit{class rank signature}, that can be represented as a \textit{class likelihood matrix} (CLM) with rows corresponding to predicted ID classes and columns to their ranks. Each matrix element signifies the likelihood of a particular ID class occurring at a specific rank. This likelihood is computed by analysing predicted class rankings across training samples. 

In Figure~\ref{fig:AUROC_figure}, we show the consistent performance exhibited by ExCeL, compared to existing post-hoc baselines in terms of the mean overall rank. To summarise, we make the following contributions.

\begin{itemize}
    \item We show that the collective information spanning all classes and training samples, embedded in the output layer of trained DNNs through predicted class ranks, can be effectively used to improve OOD detection. Consequently, we emphasise the existence of a \textit{class rank signature} for each ID class, frequently evident in ID data but not in OOD data. 

    \item We represent the class rank signature as a two-dimensional \textit{class likelihood matrix} and propose a novel post-hoc OOD detection score named ExCeL, that combines the extreme information provided by the max logit and the collective information provided by the class likelihood matrix.  
    
    \item We validate ExCeL through extensive experiments conducted in the OpenOOD benchmarking library. Compared to twenty-one existing post-hoc baselines, our method consistently ranks among the top five methods in overall near OOD and far OOD detection on CIFAR100 and ImageNet-200. Furthermore, ExCeL achieves the top mean overall rank on both CIFAR100 and ImageNet-200 outperforming all other baselines -- that are working well in one dataset but not in the other.
\end{itemize}

The rest of the paper is organised as follows. In Section~\ref{sec:related_work}, we present the related work, while Section~\ref{sec:background} provides the background and the preliminaries related to OOD detection. We provide an overview of our methodology in Section~\ref{sec:methodology}, followed by a detailed explanation of our experiment setups in Section~\ref{sec:experiments}. Next, we present the results of our experiments in Section~\ref{sec:results}, together with an analysis of the findings and outcomes. Finally, Section~\ref{sec:conclusion} concludes the paper.

\section{Related Work}
\label{sec:related_work}

A plethora of recent work attempted to address the challenge of OOD detection~\cite{yang2021generalized}. These techniques can be broadly categorised into three main groups: \textit{post-hoc inference methods}, \textit{training methods without outlier data}, and \textit{training methods with outlier data}~\cite{openood}.

\subsection{Post-hoc inference methods}

Post-hoc inference methods~\cite{openmax,msp,tempscale,odin,mds,rmds,gram,ebo,opengan,gradnorm,react,mls,vim,knn,dice,rankfeat,ash,she} utilise post-processors applied to the base classifier. These works formulate an \textit{OOD score}, which, in turn, is employed to produce a binary ID/OOD prediction through thresholding. These methods are active during the inference phase and generally assume that the classifier has been trained using the standard cross-entropy loss. They extract information from either intermediate layers~\cite{gram, nmd, gradnorm} or the output layer~\cite{msp, mls, ebo, mds, rmds} of a DNN to establish the OOD score. Given that high-level semantics are more effectively captured in the output layer, much attention has been directed towards exploiting output layer features for OOD detection. Early work by Hendrycks et al.~\cite{msp} proposed the maximum softmax probability (MSP) as a reliable baseline for detecting OOD inputs. Building upon this, later studies adopted a similar approach, leveraging more extreme information from the output layer. For instance, subsequent work by Hendricks et al.~\cite{mls} used the maximum logit directly, whereas Liu et al.~\cite{ebo} employed the energy score, which is a smooth approximation for the maximum logit. 

Expanding on these strategies, some studies incorporated advanced post-processing techniques~\cite{odin, dice, react} to elevate the performance exhibited by MSP and energy scores. For example, Sun et al.~\cite{react} introduced ReAct by rectifying activations at an upper limit, obtaining a modified logit vector with improved OOD separability. While these methods focus on extreme information, others explored a more comprehensive view of the information provided by the output layer~\cite{mls, knn, mds, rmds}. More specifically, Hendrycks et al.~\cite{mls} employed KL divergence between the softmax prediction vector and a reference vector to define an OOD score, considering predictions for all classes. In contrast, Sun et al.~\cite{knn} leveraged information across training samples, defining an OOD score based on the distance to the $k^{\text{th}}$ nearest neighbor. However, none of the prior works utilised collective information embedded within the output layer across all classes and training samples.

\subsection{Training methods without outlier data}
\label{subsec:train_no_outlier}
These methods incorporate regularisation techniques during training without relying on auxiliary OOD data, often referred to as outliers~\cite{confbranch,rotpred,godin,csi,arpl,mos,vos,logitnorm,cider,npos}. They include a diverse range of approaches, such as constraining vector norms~\cite{logitnorm}, modifying the decision boundary~\cite{mos}, and applying sophisticated learning methods~\cite{rotpred}. For instance, Wei et al.~\cite{logitnorm} enforced a constant vector norm on the logits to prevent their continuous increase throughout the model training. Furthermore, Huang et al.~\cite{mos} proposed to simplify the decision boundary between ID and OOD by decomposing the large semantic space into smaller groups with similar concepts. While the majority of these techniques followed a supervised learning approach, other work~\cite{rotpred,rodd} adopted self-supervised learning for OOD detection. 

\subsection{Training methods with outlier data}

In contrast to the methods discussed in Section~\ref{subsec:train_no_outlier}, these techniques harness the knowledge derived from auxiliary OOD data during model training~\cite{oe,mcd,udg,mixoe}. This allows OOD detectors to generalise well to unseen data and detect OOD inputs more effectively at test time. Within these approaches, some merely incorporate a set of outliers, while others attempt to mine of the most informative outliers, a process known as outlier mining~\cite{atom,poem}. Generally, these methods outperform post-hoc and training-based approaches without outlier data, as they expose the model to OOD characteristics to some extent during the training phase. Nonetheless, they have limitations in generalisation since the model gets exposed to only certain types of OODs.

Overall, post-hoc inference methods emerge as the standout choice for OOD detection, owing to their ease of implementation and competitive performance. \textit{While existing post-hoc detectors predominantly concentrate on either extreme or collective information, we propose ExCeL that combines both aspects available within the output layer.}

\section{Background}
\label{sec:background}

In classification tasks, the problem of out-of-distribution detection can be defined using the following setup. Let $\mathcal{X}=\mathbb{R}^d$ be the input space and $\mathcal{Y}=\{1,2,...,C\}$ be the output space. Assume a deep neural network $f : \mathcal{X} \rightarrow \mathbb{R}^{|\mathcal{Y}|}$ is trained on a set of data $D=\{(x_i, y_i)\}_{i=1}^{N}$ drawn from a distribution $\mathcal{P}$ defined on $\mathcal{X}\times\mathcal{Y}$. The network outputs a logit vector which
is used to predict the label of an input sample. Furthermore, let $\mathcal{D}_{\text{in}}$ denote the marginal distribution of $\mathcal{P}$ for $\mathcal{X}$, which represents the distribution of ID data. At test time, the model may encounter inputs from other distributions, denoted as $\mathcal{D}_{\text{out}}$, that differ from $\mathcal{D}_{\text{in}}$, and are recognised as out-of-distribution. Thus, the goal of OOD detection is to define a decision function $g$ such that for a given test input $x \in \mathcal{X}$ : 
\begin{equation}
\label{eq:ood_problem}
    g(x;f) = 
    \begin{cases}
        1 &  \text{if } x \sim \mathcal{D}_{\text{in}} \\
        0 & \text{if } x \sim \mathcal{D}_{\text{out}} \\
    \end{cases}
\end{equation}

Post-hoc detectors modify the OOD detection problem in Equation~\ref{eq:ood_problem} by leveraging a scoring function $S(x)$ and make the decision via a threshold ($\lambda$) comparison as follows.
\begin{equation}
     g_{\lambda}(x) = 
    \begin{cases}
        \text{in} &  \text{if } S(x;f) \geq \lambda \\
        \text{out} & \text{if } S(x;f) < \lambda \\
    \end{cases}
\end{equation}

\section{Methodology}
\label{sec:methodology}

\begin{figure}[t!]
    \centering
    \includegraphics[width=0.47\textwidth]{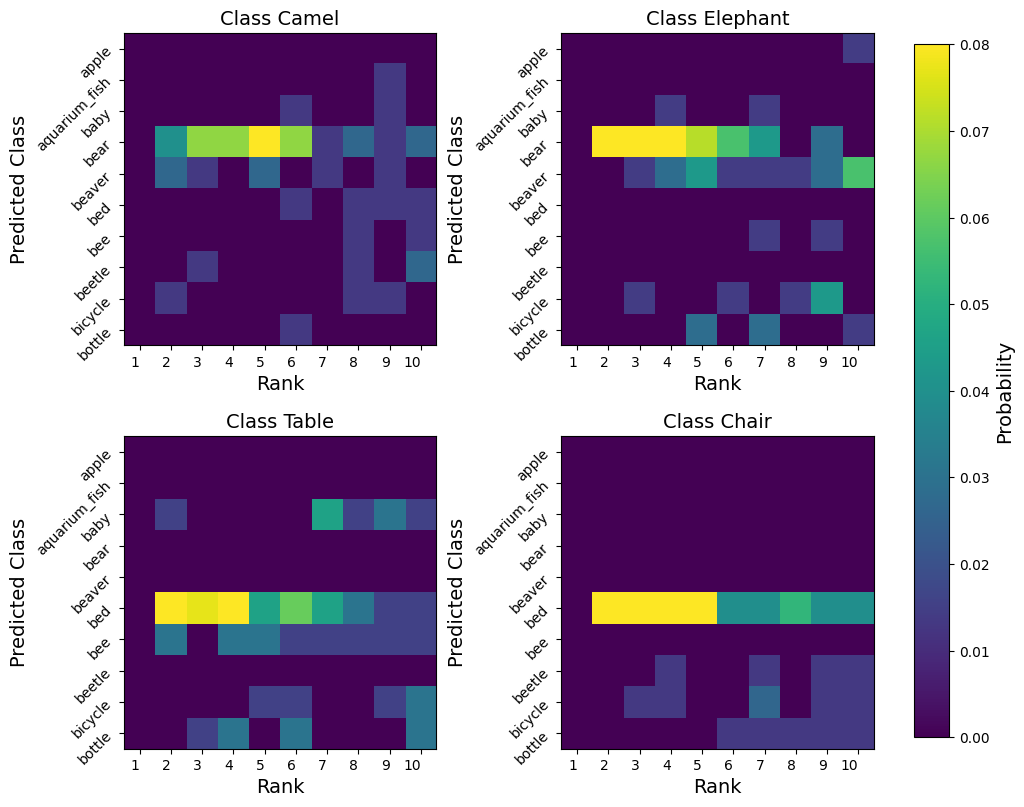}
    \caption{Class rank signatures for the top ten ranks for four base ID classes in CIFAR100. Specifically, for the base classes \textit{Camel} and \textit{Elephant}, there is a high likelihood of class \textit{Bear} appearing among the top five ranks. Similarly, for the base classes \textit{Chair} and \textit{Table}, class \textit{Bed} is observed to have a high likelihood of ranking within the top five positions. This is the central concept employed in ExCeL for OOD detection.}
    \label{fig:PPM_figure}
\end{figure}

In this section, we first explain the intuition behind our proposed method. Following that, we outline the ExCeL score computation algorithm. Lastly, we analytically justify our motivation.

\subsection{Intuition}

As discussed in Section~\ref{sec:introduction}, our idea is motivated by the observation that, during inference, when an input is predicted as a specific ID class, the rankings of the subsequent classes are more deterministic for ID data compared to OOD data. To illustrate this more clearly, we depict the likelihood of a subset of ID classes ranking among the top ten for four base classes of the CIFAR100 dataset in Figure~\ref{fig:PPM_figure}. It is important to note that the first rank is always assigned to the respective base class. We notice that when a test input is predicted as either an \textit{elephant} or a \textit{camel}, there is a strong likelihood of the class \textit{bear} appearing within the top five predictions, given its semantic proximity to these classes. Similarly, in the case of an input predicted as a \textit{chair} or a \textit{table}, the class \textit{bed} is highly likely to be among the top five predictions. We refer to these distinct patterns associated with each class as \textit{class rank signatures}. Notably, such trends are absent in OOD data, enabling us to leverage this information to design ExCeL based on the predicted class ranking for efficient OOD detection.

\subsection{ExCeL score computation}

The ExCeL score computation comprises four steps; two steps involving pre-computation using training samples and two steps executed at test time. Firstly, we calculate the \textit{class likelihood matrix} for each ID class by leveraging correctly classified training samples specific to that class. Following this, the class likelihood matrix undergoes a smoothing process to amplify the influence of frequently occurring classes while penalising less prevalent ones. During test time, based on the top predicted class of an input, a \textit{rank score} is computed using the relevant class likelihood matrix. This score captures collective information from the ranking of predicted classes. Finally, the rank score is linearly combined with the max logit to compute the final ExCeL score for OOD detection. These steps are further discussed in the following sections.

\subsubsection{Generating the Class Likelihood Matrix}
\label{subsubsec:CLM}

The objective of the class likelihood matrix (CLM) is to model the probability mass function (PMF) across all ID classes within each rank. To achieve this, we start by filtering correctly classified training samples for each specific class, and rank the remaining $C-1$ classes based on their corresponding logit values. Within each rank, we then calculate the likelihood of a particular class across the training samples. In Figure~\ref{fig:PPM_figure}, when an input is predicted as a \textit{chair}, the probability of class \textit{bed} appearing in one of the next four ranks (i.e., ranks 2-5) is approximately 0.08. Similarly, when an input is predicted as \textit{elephant}, class \textit{bear} appears in ranks 2-4, with a probability of approximately 0.08, in each position.

Likewise, for each ID class $c$, an element $p_{ij}^c$ in the class likelihood matrix ($P_{c}\in \mathbb{R}^{C\times C}$) indicates the probability of class $i$ occurring in the $j^{\text{th}}$ rank when an input is predicted as class $c$ as shown in Equation~\ref{eq:clm}. 
\begin{equation}
\label{eq:clm}
P_c = \begin{pmatrix}
    p_{11}^c & p_{12}^c & \dots & p_{1C}^c \\
    p_{21}^c & p_{22}^c & \dots & p_{2C}^c \\
    \vdots & \vdots & \ddots & \vdots \\
    p_{C1}^c & p_{C2}^c & \dots & p_{CC}^c \\
\end{pmatrix},
\quad p_{ij}^c = \frac{n_{ij}^c}{N_c}
\end{equation} 
Here, $n_{ij}^c$ represents the number of occurrences where class $i$ appears at rank $j$ among correctly classified samples in class $c$. Furthermore, $N_c$ denotes the total number of correctly classified samples in class $c$, and $C$ represents the total number of ID classes. It is important to highlight that when calculating the likelihood matrix for any class $c$, we exclusively take into account the correctly classified training samples belonging to that class. Therefore, the top rank in the likelihood matrix is invariably occupied by the class $c$ itself.
\begin{equation}
\label{eq:first_column}
    p_{i1}^c = \begin{cases}
        1  &  \text{if } i=c \\
        0  & \text{otherwise}\\
    \end{cases}. 
\end{equation}

\subsubsection{Smoothing the CLM}
\label{subsec:smoothing}
DNNs tend to exhibit some degree of overfitting to the training set. Therefore, some samples may lose the correlation between similar classes during training. This will induce noise in the CLM computed in Section~\ref{subsubsec:CLM}. Hence, to extract high-level information from the CLM, we incorporate a smoothing step, employing a piecewise function based on the following criteria:
\begin{itemize}
    \item For classes frequently occurring in a specific rank, a fixed high reward is assigned.
    \item If the likelihood of a class, though not highly significant, surpasses that of a random prediction, a small reward is given.
    \item If the likelihood of a class is worse than a random prediction but not zero, a small penalty is imposed.
    \item Classes that do not appear in a specific rank receive a fixed high penalty.
\end{itemize}
Specifically, let $\hat{P}_c$ be the smoothed likelihood matrix of $P_c$. Each element $\hat{p}_{ij}^c$ in $\hat{P}_c$ is determined based on the corresponding value $p_{ij}^c$ in $P_c$ based on Equation~\ref{eq:smoothing}. 
\begin{equation}
    \label{eq:smoothing}
    \hat{p}_{ij}^c = \begin{cases}
               \frac{a}{C-1}  &  \text{if } p_{ij}^c \geq \frac{b}{C-1} \\
               \frac{1}{C-1}  &  \text{if } \frac{1}{C-1} \leq p_{ij}^c < \frac{b}{C-1} \\
               -\frac{1}{C-1} &  \text{if } 0 < p_{ij}^c < \frac{1}{C-1} \\
               -\frac{a}{C-1}  &  \text{if } p_{ij}^c=0 \\
    \end{cases}
\end{equation}
Here, $a$ corresponds to the reward, while $b$ denotes the high likelihood threshold. We use the validation set to determine these hyperparameters. The smoothed class likelihood matrix is then employed to calculate the rank score that provides a measure of how closely a prediction aligns with the distinct class rank signature.

\subsubsection{Computing the rank score}

For a given test image $x$, let the predicted class ranking be [$c_1$, $c_2$, $c_3$,..., $c_C$], where $c_1$ and $c_C$ are the classes with the highest and the lowest logit values respectively. Since the top predicted class is $c_1$, we use the smoothed class likelihood matrix of class $c_1$, denoted by $\hat{P}_{c_1}$ for the rank score calculation. Thus, we compute the rank score ($\mathcal{RS}$) of $x$ as, 
\begin{equation}
\label{eq:rank_score}
    \mathcal{RS}(x) = \sum_{i=1}^{C} \hat{p}_{c_ii}^{c_1}
\end{equation}
Since we consider the CLM associated with the top-ranked class when computing the rank score, it is worth noting that the first term (i.e., $\hat{p}_{c_11}^{c_1}$ ) consistently yields 1 (i.e., analogous to $\frac{a}{C-1}$ in $\hat{P}_c$) for all inputs, in accordance with Equation~\ref{eq:first_column}.  Consequently, the presence of the first term in Equation~\ref{eq:rank_score} merely introduces a constant shift to the score, without actively contributing to the discrimination between ID and OOD. However, for completeness, we retain the first term in rank score computation.

Moreover, the rank score can also be computed efficiently via matrix operations. In order to achieve this, the predicted ranking is represented as a one-hot encoded matrix ($\rho$) with the predicted classes as the rows and ranks as columns. For example, in a four-class classification problem, if the predicted class ranking for an input $\hat{x}$ is [$1,4,2,3$], $\rho_{\hat{x}}$ would be as follows. 
\begin{equation}
\rho_{\hat{x}} = \begin{pmatrix}
    1 & 0 & 0 & 0 \\
    0 & 0 & 1 & 0 \\
    0 & 0 & 0 & 1 \\
    0 & 1 & 0 & 0 \\
\end{pmatrix}
\end{equation}
Accordingly, for an input $x$, if the one-hot encoded predicted class ranking matrix is $\rho_x$, we can compute the rank score as,
\begin{equation}
    \mathcal{RS}(x) = \mathrm{tr}[(\hat{P}_c)^T\rho_x]
\end{equation}

A higher rank score is indicative of accumulating more rewards as per Equation~\ref{eq:smoothing}. This suggests a strong alignment between the predicted class ranking and the rank patterns observed in training samples, indicating that the input is highly likely to be ID. In this way, the rank score encompasses the collective information spanning all ID classes and training samples, which is then utilised to improve OOD detection.

\subsubsection{Combining with the maximum logit}

The rank score draws on collective information from subsequent ranks, excluding the top rank, which itself holds valuable information inherent to ID data. Therefore, as the final step, we combine the rank score with the logit value of the top predicted class, referred to as \textit{MaxLogit} by Hendrycks et al.~\cite{mls}, to compute the \textit{ExCeL} score for OOD detection. Since MaxLogit contains extreme information within the output layer, the final \textit{ExCeL} score incorporates both extreme and collective information embedded in the output layer. We define the ExCeL score as a linear combination of the rank score and the MaxLogit as per Equation~\ref{eq:excel}.
\begin{equation}
\label{eq:excel}
    \text{ExCeL}(x) = \alpha \cdot \mathcal{RS}(x) + (1-\alpha) \cdot \text{MaxLogit}
\end{equation}
Here, $\alpha$ balances the trade-off between using collective and extreme information for OOD detection. We fine-tune $\alpha$ using a validation set following the same approach used in Section~\ref{subsec:smoothing} to fine-tune $a$ and $b$.

\subsection{Analytical justification}
\label{subsec:justification}
We next analytically justify the existence of distinct patterns in the class ranking that can be exploited to improve OOD detection. Suppose for any predicted ID class, the remaining $C-1$ classes occur uniformly distributed across the subsequent ranks. Then, for any class $c$, the likelihood matrix, denoted as Equation~\ref{eq:clm}, takes the following form:
\begin{equation}
    p_{ij}^c = \begin{cases}
         1  &  \text{if } (i=c \text{ and } j=1)  \\
         0  &  \text{if } (i\neq c \text{ and } j=1) \text{ or } (i=c \text{ and } j\neq 1) \\
        \frac{1}{C-1} &  \text{otherwise} \\
    \end{cases}.
\end{equation}
Thus, for any predicted class ranking of an input $x$, the rank score would be,
\begin{equation}
\begin{aligned}
       \mathcal{RS}(x) &=  \sum_{i=1}^{C} \hat{p}_{c_ii}^{c} \\
       &= \frac{a}{C-1} + \frac{1}{C-1} + ... + \frac{1}{C-1} \\
       &= \frac{a}{C-1} + 1 = k \text{ (constant)}
\end{aligned}.
\end{equation}
Subsequently, we can compute the ExCeL score as,  
\begin{equation}
\label{eq:excel_just}
    \begin{aligned}
        \text{ExCeL}(x) &= \alpha \cdot \mathcal{RS}(x) + (1-\alpha) \cdot \text{MaxLogit} \\
        &= \alpha \cdot k + (1-\alpha) \cdot \text{MaxLogit} \\
    \end{aligned}.
\end{equation}
As shown in Equation~\ref{eq:excel_just}, when classes appear uniformly at random, the rank score remains constant, leading the ExCeL score to correspond to a linearly transformed MaxLogit. In this case, the OOD detection performance of ExCeL score would be identical to that of the MaxLogit, since a linear transformation would not impact the separability between ID and OOD. Hence, if ExCeL demonstrates enhanced OOD detection performance compared to MaxLogit, it would affirm the presence of unique class rank signatures within ID classes.
\section{Experiments}
\label{sec:experiments}

We evaluate ExCeL over common OOD detection benchmarks. To ensure a fair comparison with various baselines, we use the OpenOOD\footnote{https://github.com/Jingkang50/OpenOOD.} library by Zhang et al.~\cite{openood}. We have implemented ExCeL in the OpenOOD environment and the code will be made publicly available upon the acceptance of the manuscript.

\subsection{Datasets}
\label{Sec:Datasets}

We use CIFAR100~\cite{cifar100} and ImageNet-200 (a.k.a., TinyImageNet)~\cite{tin} as ID data in our experiments. Each ID dataset is evaluated against near-OOD and far-OOD datasets. As near-OOD data follow a closer distribution to ID compared to far-OOD, near-OOD detection is more challenging than far-OOD detection. For CIFAR100, CIFAR10~\cite{cifar100} and TinyImageNet datasets serve as near-OOD, while MNIST~\cite{mnist}, SVHN~\cite{svhn}, Textures~\cite{textures}, and Places365~\cite{places} are considered as far-OOD. Similarly, for TinyImageNet, SSB-hard~\cite{ssb_hard} and NINCO~\cite{ninco} datasets are used as near-OOD, while iNaturalist~\cite{inaturalist}, Textures~\cite{textures}, and OpenImage-O~\cite{vim} datasets are used as far-OOD. For consistency, we adopt the same train, validation, and test splits used by OpenOOD developers in implementing our method. 

\subsection{Models}

For both CIFAR-100 and TinyImageNet datasets, we use ResNet-18~\cite{resnet18} as the base model. Each model is trained for 100 epochs using the standard cross-entropy loss. We use the SGD optimiser with a momentum of 0.9, a learning rate of 0.1, and a cosine annealing decay schedule~\cite{loshchilov2016sgdr}. Furthermore, we incorporate a weight decay of 0.0005, and employ batch sizes of 128 and 256 for CIFAR100 and ImageNet-200, respectively.

\subsection{Comparison with baselines}

We compare ExCeL with twenty-one existing post-hoc inference methods provided by the OpenOOD library. These baselines include early OOD detection methods such as maximum softmax probability (MSP)~\cite{msp}, Mahalanobis distance (MDS)~\cite{mds}, and OpenMax~\cite{openmax}, as well as state-of-the-art approaches like SHE~\cite{she}, ASH~\cite{ash}, and DICE~\cite{dice}.

\subsection{Evaluation metrics}

We employ two metrics to evaluate the OOD detection performance: i) FPR95, which measures the false positive rate of OOD samples when the true positive rate of ID samples is at 95\%; ii) AUROC, representing the area under the receiver operating curve. An effective OOD detector will exhibit a low FPR95 alongside a high AUROC. We also measure the overall performance of each method by computing the mean of near and far-OOD performances. Finally, to compare the performance of a method with other baselines across ID datasets, we define the \textit{mean overall rank} which is computed as the average of overall ranks for AUROC and FPR95 as per Equation~\ref{eq:mean_rank}.      
\begin{equation}
\label{eq:mean_rank}
\text{Mean Overall Rank} = \frac{\mathcal{R}_{\text{overall}}^{\text{AUROC}} + \mathcal{R}_{\text{overall}}^{\text{FPR95}}}{2}    
\end{equation}
Moreover, we report the mean and the standard deviation of the above metrics computed over three independent runs in Section~\ref{sec:results}.

\begin{table*}[t!]
\centering
\footnotesize
\caption{\centering Comparison of post-hoc OOD detectors for CIFAR100 (ID). The performance rank of each method is indicated within brackets. Top five values are marked in \textbf{bold}. \break Results on CIFAR100 indicate that the \textbf{ExCeL} and \textbf{RMDS} methods share the best performance, with a mean overall rank of 1.5. The top five ranks also include ReAct, GEN, and MaxLogit.}
\label{tab:cifar100}
\begin{tabular}{lccc|ccc|c}
\hline
\multirow{2}{*}{Post-processor} & \multicolumn{3}{c|}{AUROC (\%) $\uparrow$} & \multicolumn{3}{c|}{FPR95 (\%) $\downarrow$} & \multirow{2}{*}{Mean Overall Rank} \\
\cline{2-7}
& Near-OOD & Far-OOD & Overall & Near-OOD & Far-OOD & Overall & \\
\hline
OpenMax~\cite{openmax} & 76.41 $\pm$ 0.25 (15) & 79.48 $\pm$ 0.41 (11) & 77.95 (14) & 56.58 $\pm$ 0.73 (9) & 54.50 $\pm$ 0.68 (6) & \textbf{55.54} (4) & 9.0 \\
MSP~\cite{msp} & 80.27 $\pm$ 0.11 (7) & 77.76 $\pm$ 0.44 (14) & 79.02 (12) & \textbf{54.80 $\pm$ 0.33} (3) & 58.70 $\pm$ 1.06 (12) & 56.75 (10) & 11.0 \\
TempScale~\cite{tempscale} & \textbf{80.90 $\pm$ 0.07} (4) & 78.74 $\pm$ 0.51 (13) & 79.82 (8) & \textbf{54.49 $\pm$ 0.48} (2) & 57.94 $\pm$ 1.14 (11) & 56.22 (8) & 8.0 \\
ODIN~\cite{odin} & 79.90 $\pm$ 0.11 (10) & 79.28 $\pm$ 0.21 (12) & 79.59 (10) & 57.91 $\pm$ 0.51 (10) & 58.86 $\pm$ 0.79 (13) & 58.39 (13) & 11.5 \\
MDS~\cite{mds} & 58.69 $\pm$ 0.09 (20) & 69.39 $\pm$ 1.39 (18) & 64.04 (20) & 83.53 $\pm$ 0.60 (19) & 72.26 $\pm$ 1.56 (21) & 77.90 (19) & 19.5 \\
MDSEns~\cite{mds} & 46.31 $\pm$ 0.24 (22) & 66.00 $\pm$ 0.69 (22) & 56.16  (22) & 95.88 $\pm$ 0.04 (22) & 66.74 $\pm$ 1.04 (17) & 81.31 (21) & 21.5 \\
RMDS~\cite{rmds} & 80.15 $\pm$ 0.11 (9) & \textbf{82.92 $\pm$ 0.42} (1) & \textbf{81.54} (1) & \textbf{55.46 $\pm$ 0.41} (5) & \textbf{52.81 $\pm$ 0.63} (3) & \textbf{54.14} (2) & \textbf{1.5} \\
Gram~\cite{gram} & 51.66 $\pm$ 0.77 (21) & 73.36 $\pm$ 1.08 (17) & 62.51 (21) & 92.28 $\pm$ 0.29 (21) & 64.44 $\pm$ 2.37 (16) & 78.36 (20) & 20.5 \\
EBO~\cite{ebo} & \textbf{80.91 $\pm$ 0.08} (3) & 79.77 $\pm$ 0.61 (8) & 80.34 (7) & 55.62 $\pm$ 0.61 (7) & 56.59 $\pm$ 1.38 (8) & 56.11 (7) & 7.0 \\
OpenGAN~\cite{opengan} & 65.98 $\pm$ 1.26 (18) & 67.88 $\pm$ 7.16 (20) & 66.93 (18) & 76.52 $\pm$ 2.59 (16) & 70.49 $\pm$ 7.38 (19) & 73.51 (16) & 17.0 \\
GradNorm~\cite{gradnorm} & 70.13 $\pm$ 0.47 (17) & 69.14 $\pm$ 1.05 (19) & 69.64 (17) & 85.58 $\pm$ 0.46 (20) & 83.68 $\pm$ 1.92 (22) & 84.63 (22) & 19.5 \\
ReAct~\cite{react} & \textbf{80.77 $\pm$ 0.05} (5) & 80.39 $\pm$ 0.49 (6) & \textbf{80.58} (4) & 56.39 $\pm$ 0.34 (8) & \textbf{54.20 $\pm$ 1.56} (5) & \textbf{55.30} (3) & \textbf{3.5} \\
KLM~\cite{mls} & 76.56 $\pm$ 0.25 (14) & 76.24 $\pm$ 0.52 (16) & 76.40 (16) & 77.92 $\pm$ 1.31 (17) & 71.65 $\pm$ 2.01 (20) & 74.79 (17) & 16.5 \\
VIM~\cite{vim} & 74.98 $\pm$ 0.13 (16) & \textbf{81.70 $\pm$ 0.62} (4) & 78.34 (13) & 62.63 $\pm$ 0.27 (14) & \textbf{50.74 $\pm$ 1.00} (1) & 56.69 (9) & 11.0 \\
KNN~\cite{knn} & 80.18 $\pm$ 0.15 (8) & \textbf{82.40 $\pm$ 0.17} (2) & \textbf{81.29} (3) & 61.22 $\pm$ 0.14 (13) & \textbf{53.65 $\pm$ 0.28} (4) & 57.44 (12) & 7.5 \\
DICE~\cite{dice} & 79.38 $\pm$ 0.23 (11) & 80.01 $\pm$ 0.18 (7) & 79.70 (9) & 57.95 $\pm$ 0.53 (11) & 56.25 $\pm$ 0.60 (7) & 57.10 (11) & 10.0 \\
RankFeat~\cite{rankfeat} & 61.88 $\pm$ 1.28 (19) & 67.10 $\pm$ 1.42 (21) & 64.49 (19) & 80.59 $\pm$ 1.10 (18) & 69.45 $\pm$ 1.01 (18) & 75.02 (18) & 18.5 \\
ASH~\cite{ash} & 78.20 $\pm$ 0.15 (13) & \textbf{80.58 $\pm$ 0.66} (5) & 79.39 (11) & 65.71 $\pm$ 0.24 (15) & 59.20 $\pm$ 2.46 (14) & 62.46 (15) & 13.0 \\
SHE~\cite{she} & 78.95 $\pm$ 0.18 (12) & 76.92 $\pm$ 1.16 (15) & 77.94 (15) & 59.07 $\pm$ 0.25 (12) & 64.12 $\pm$ 2.70 (15) & 61.60 (14) & 14.5 \\
GEN~\cite{gen} & \textbf{81.31 $\pm$ 0.08} (1) & 79.68 $\pm$ 0.75 (9) & \textbf{80.50} (5) & \textbf{54.42 $\pm$ 0.33} (1) & 56.71 $\pm$ 1.59 (9) & \textbf{55.57} (5) & \textbf{5.0} \\
MaxLogit~\cite{mls} & \textbf{81.05 $\pm$ 0.07} (2) & 79.67 $\pm$ 0.57 (10) & 80.36 (6) & 55.47 $\pm$ 0.66 (6) & 56.73 $\pm$ 1.33 (10) & 56.10 (6) & \textbf{6.0}  \\
\hline
ExCeL (Ours) & 80.70 $\pm$ 0.06 (6) & \textbf{82.04 $\pm$ 0.90} (3) & \textbf{81.37} (2) & \textbf{55.21 $\pm$ 0.56} (4) & \textbf{52.24 $\pm$ 1.90} (2) & \textbf{53.73} (1) & \textbf{1.5} \\
\hline
\end{tabular}
\end{table*}

\subsection{Hyperparameter tuning}

Fine-tuning hyperparameters on a validation set is widely adopted in prior OOD detection work~\cite{oe,knn,ebo}. Following a similar approach, we determine the three parameters associated with ExCeL using the validation set. By performing a grid search on $a$, $b$, and $\alpha$, we discovered the best hyperparameter combination for both CIFAR100 and ImageNet-200 datasets, is $a=10$, $b=5$, and $\alpha=0.8$.

\begin{table*}[h!]
\centering
\footnotesize
\caption{\centering Comparison of post-hoc OOD detectors for ImageNet-200 (ID). The performance rank of each method is indicated within brackets. Top five values are marked in \textbf{bold}. \break Results on ImageNet-200 indicate that the \textbf{ExCeL} and \textbf{GEN} methods share the best performance, with a mean overall rank of 3.0. The top five ranks also include KNN, ASH, and TempScale.}
\label{tab:tin}
\begin{tabular}{lccc|ccc|c}
\hline
\multirow{2}{*}{Post-processor} & \multicolumn{3}{c|}{AUROC (\%) $\uparrow$} & \multicolumn{3}{c|}{FPR95 (\%) $\downarrow$} & \multirow{2}{*}{Mean Overall Rank} \\
\cline{2-7}
& Near-OOD & Far-OOD & Overall & Near-OOD & Far-OOD & Overall & \\
\hline
 OpenMax~\cite{openmax} & 80.27 $\pm$ 0.10 (13) & 90.20 $\pm$ 0.17 (12) & 85.24 (13) & 63.48 $\pm$ 0.25 (12) & 33.12 $\pm$ 0.66 (8) & 48.30 (12) & 12.5 \\
    MSP~\cite{msp} & \textbf{83.34 $\pm$ 0.06} (3) & 90.13 $\pm$ 0.09 (13) & 86.74 (8) & \textbf{54.82 $\pm$ 0.35} (2) & 35.43 $\pm$ 0.38 (13) & 45.13 (7) & 7.5 \\
    TempScale~\cite{tempscale} & \textbf{83.69 $\pm$ 0.04} (1) & 90.82 $\pm$ 0.09 (10) & \textbf{87.26} (4) & \textbf{54.82 $\pm$ 0.23} (2) & 34.00 $\pm$ 0.37 (9) & 44.41 (6) & \textbf{5.0} \\
    ODIN~\cite{odin} & 80.27 $\pm$ 0.08 (13) & \textbf{91.71 $\pm$ 0.19} (5) & 85.99 (11) & 66.76 $\pm$ 0.26 (14) & 34.23 $\pm$ 1.05 (11) & 50.50 (14) & 12.5 \\
    MDS~\cite{mds} & 61.93 $\pm$ 0.51 (19) & 74.72 $\pm$ 0.26 (18) & 68.33 (19) & 79.11 $\pm$ 0.31 (17) & 61.66 $\pm$ 0.27 (17) & 70.39 (17) & 18.0 \\
    MDSEns~\cite{mds} & 54.32 $\pm$ 0.24 (22) & 69.27 $\pm$ 0.57 (21) & 61.80 (21) & 91.75 $\pm$ 0.10 (21) & 80.96 $\pm$ 0.38 (20) & 86.36 (21) & 21.0 \\
    RMDS~\cite{rmds} & \textbf{82.57 $\pm$ 0.25} (5) & 88.06 $\pm$ 0.34 (16) & 85.32 (12) & \textbf{54.02 $\pm$ 0.58} (1) & 32.45 $\pm$ 0.79 (7) & \textbf{43.24} (3) & 7.5 \\
    Gram~\cite{gram} & 67.67 $\pm$ 1.07 (18) & 71.19 $\pm$ 0.24 (20) & 69.43 (18) & 86.40 $\pm$ 1.21 (20) & 84.36 $\pm$ 0.78 (21) & 85.38 (20) & 19.0 \\
    EBO~\cite{ebo} & 82.50 $\pm$ 0.05 (6) & 90.86 $\pm$ 0.21 (9) & 86.68 (9) & 60.24 $\pm$ 0.57 (9) & 34.86 $\pm$ 1.30 (12) & 47.55 (11) & 10.0 \\
    OpenGAN~\cite{opengan} & 59.79 $\pm$ 3.39 (20) & 73.15 $\pm$ 4.07 (19) & 66.47 (20) & 84.15 $\pm$ 3.85 (19) & 64.16 $\pm$ 9.33 (18) & 74.16 (18) & 19.0 \\
    GradNorm~\cite{gradnorm} & 72.75 $\pm$ 0.48 (17) & 84.26 $\pm$ 0.87 (17) & 78.51 (17) & 82.67 $\pm$ 0.30 (18) & 66.45 $\pm$ 0.22 (19) & 74.56 (19) & 18.0 \\
    ReAct~\cite{react}& 81.87 $\pm$ 0.98 (9) & \textbf{92.31 $\pm$ 0.56} (3) & 87.09 (6) & 62.49 $\pm$ 2.19 (11) & \textbf{28.50 $\pm$ 0.95} (5) & 45.50 (8) & 7.0 \\
    KLM~\cite{mls} & 80.76 $\pm$ 0.08 (12) & 88.53 $\pm$ 0.11 (15) & 84.65 (16) & 70.26 $\pm$ 0.64 (16) & 40.90 $\pm$ 1.08 (15) & 55.58 (16) & 16.0 \\
    VIM~\cite{vim} & 78.68 $\pm$ 0.24 (16) & 91.26 $\pm$ 0.19 (7) & 84.97 (15) & 59.19 $\pm$ 0.71 (6) & \textbf{27.20 $\pm$ 0.30} (1) & \textbf{43.20} (2) & 8.5 \\
    KNN~\cite{knn} & 81.57 $\pm$ 0.17 (11) & \textbf{93.16 $\pm$ 0.22} (2) & \textbf{87.37} (3) & 60.18 $\pm$ 0.52 (8) & \textbf{27.27 $\pm$ 0.75} (2) & \textbf{43.73} (5) & \textbf{4.0} \\
    DICE~\cite{dice} & 81.78 $\pm$ 0.14 (10) & 90.80 $\pm$ 0.31 (11) & 86.29 (10) & 61.88 $\pm$ 0.67 (10) & 36.51 $\pm$ 1.18 (14) & 49.20 (13) & 11.5 \\
    RankFeat~\cite{rankfeat} & 56.92 $\pm$ 1.59 (21) & 38.22 $\pm$ 3.85 (22) & 47.57 (22) & 92.06 $\pm$ 0.23 (22) & 97.72 $\pm$ 0.75 (22) & 94.89 (22) & 22.0 \\
    ASH~\cite{ash} & 82.38 $\pm$ 0.19 (8) & \textbf{93.90 $\pm$ 0.27} (1) & \textbf{88.14} (1) & 64.89 $\pm$ 0.90 (13) & \textbf{27.29 $\pm$ 1.12} (3) & 46.09 (9) & \textbf{5.0} \\
    SHE~\cite{she} & 80.18 $\pm$ 0.25 (15) & 89.81 $\pm$ 0.61 (14) & 85.00 (14) & 66.80 $\pm$ 0.74 (15) & 42.17 $\pm$ 1.24 (16) & 54.49 (15) & 14.5 \\
    GEN~\cite{gen} & \textbf{83.68 $\pm$ 0.06} (2) & 91.36 $\pm$ 0.10 (6) & \textbf{87.52} (2) & \textbf{55.20 $\pm$ 0.20} (4) & 32.10 $\pm$ 0.59 (6) & \textbf{43.65} (4) & \textbf{3.0} \\
    MaxLogit~\cite{mls} & \textbf{82.90 $\pm$ 0.04} (4) & 91.11 $\pm$ 0.19 (8) & 87.01 (7) & 59.76 $\pm$ 0.59 (7) & 34.03 $\pm$ 1.21 (10) & 46.90 (10) & 8.5  \\
    \hline
    ExCeL (Ours) & 82.40 $\pm$ 0.04 (7) & \textbf{91.97 $\pm$ 0.27} (4) & \textbf{87.19} (5) & \textbf{57.90 $\pm$ 0.40} (5) & \textbf{28.45 $\pm$ 0.80} (4) & \textbf{43.18} (1) & \textbf{3.0}\\
\hline
\end{tabular}
\end{table*}
\section{Results and Analysis}
\label{sec:results}

\subsection{Results}

We present our main results in Table~\ref{tab:cifar100} and Table~\ref{tab:tin}. Note that due to space constraints, we present the average performance for both near-OOD and far-OOD across all OOD benchmarks in each group ({\bf cf.} Section~\ref{Sec:Datasets}), along with the overall OOD performance. Per-dataset statistics can be found in the appendix. According to the results, ExCeL consistently stands out among the five top-performing methods in most instances. For instance, in CIFAR100, ExCeL is ranked second in overall AUROC, first in overall FPR95, and has the equal best mean overall rank with RMDS. Similarly, in ImageNet-200, ExCeL is ranked fifth in overall AUROC and first in overall FPR95. Again, ExCeL is ranked equal first in mean overall rank, but this time, sharing it with a different method, GEN.

We can also observe from the results that most of the other baselines exhibit strong performance in specific cases but demonstrate only moderate or poor performance in others. For example, RMDS excels in the CIFAR100 ranking, sharing equal best mean overall rank with ExCeL. However, its performance lags behind in ImageNet-200. Similarly, GEN performs exceptionally well in ImageNet-200, but falls behind in CIFAR100. In contrast, ExCeL delivers consistent results in both datasets, achieving a mean overall rank of 1.5 in CIFAR100 and 3.0 in ImageNet-200.

Finally, the results also show that ExCeL performs slightly better in far-ODD detection than near-OOD detection. For example, in CIFAR100, ExCeL is ranked second and fourth in terms of FPR95 for far and near ODD, respectively. Similarly, in ImageNet-200, ExCeL is ranked fourth and fifth in terms of FPR95 for far and near ODD. This can be explained using the characteristics of likelihood matrices in different datasets.

\begin{figure}[t!]
    \centering
     \begin{subfigure}{0.22\textwidth}
        \includegraphics[width=\linewidth]{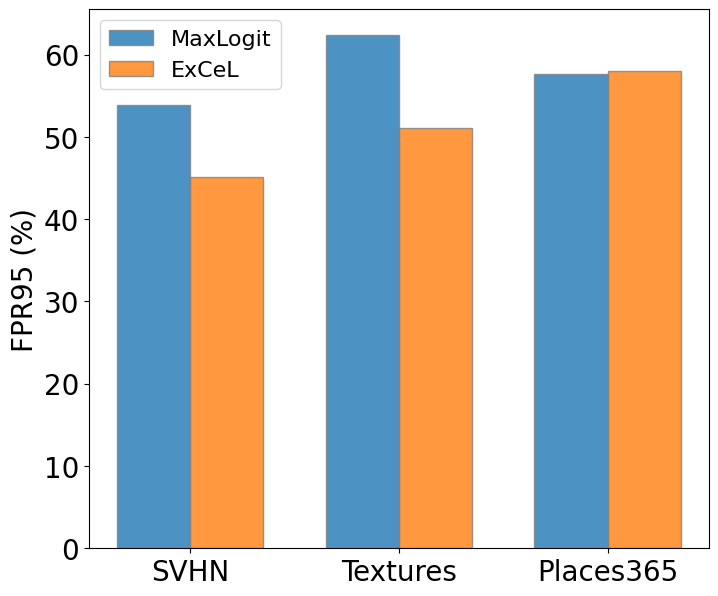}
        \caption{ID - CIFAR100}
        \label{fig:cifar100_fpr}
    \end{subfigure}
    \hfill
    \begin{subfigure}{0.22\textwidth}
        \includegraphics[width=\linewidth]{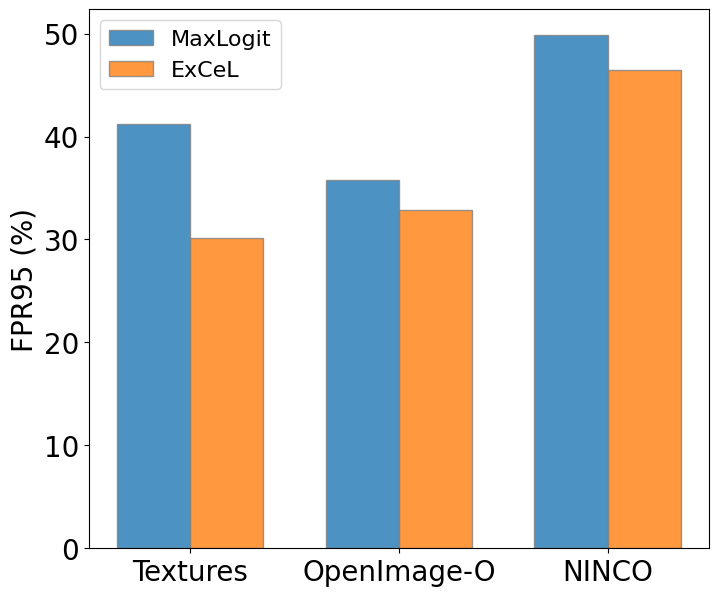}
        \caption{ID - ImageNet-200}
        \label{fig:tin_fpr}
    \end{subfigure}
    \caption{FPR95 comparison between MaxLogit and ExCeL.}
    \label{fig:fpr_comparison}
\end{figure}

\subsection{Class likelihood matrix and Maxlogit}

As analysed in Section~\ref{subsec:justification}, an improvement in the OOD detection in ExCeL compared to MaxLogit confirms the existence of unique class rank patterns in ID classes. First, to validate this, we show the FPR95 comparison of ExCeL and MaxLogit in Figure~\ref{fig:fpr_comparison} on CIFAR100 and ImageNet-200 datasets. We observe two key behaviours in the comparison. For some OOD datasets ExCeL achieves a significant improvement in FPR95 compared to MaxLogit, while in others, the performance of ExCeL remains closely aligned with MaxLogit. For example, when CIFAR100 serves as ID, ExCeL significantly improves the OOD detection in SVHN and Textures, while the performance on Places365 is similar to MaxLogit.

\begin{figure}[t!]
    \centering
    \includegraphics[width=0.51\textwidth]{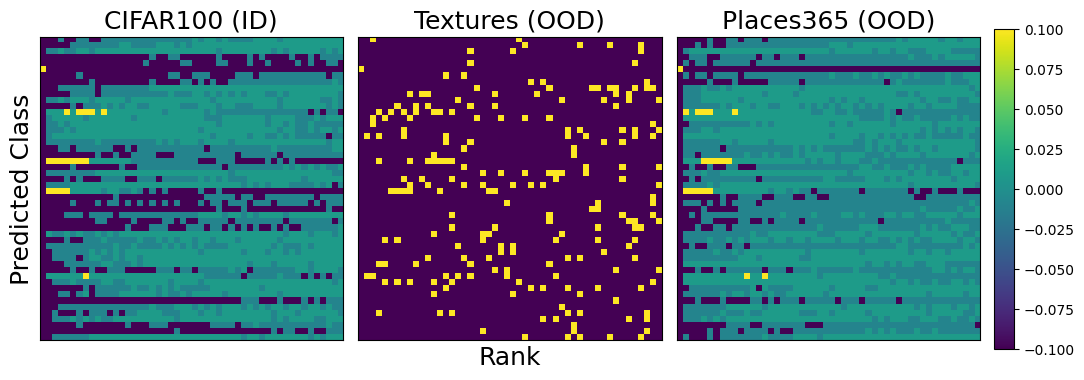}
    \caption{Class likelihood matrices computed for ID and OOD samples predicted as a selected ID class in CIFAR100. We use Textures and Places365 as OOD samples. We can see clear clusters of classes occurring mainly within the top ranks for ID data. In Textures, the predicted class rankings show a random and sparse behaviour, while in Places365, the occurrence of classes is close to a uniformly random distribution.}
    \label{fig:heat_maps}
\end{figure}

The reason for this can be explained using the class likelihood matrices of ID and OOD data as shown in Figure~\ref{fig:heat_maps}. Here, we show how the subsequent classes are ranked in ID and OOD samples for a selected class in CIFAR100. Specifically, for CIFAR100, we see clear clusters of classes occurring mainly within the top ranks for ID data, indicating a unique class rank signature. In contrast, the class occurrence in Textures data looks random and sparse which allows the ExCeL score to separate the two datasets effectively. On the other hand, the occurrence of classes in Places365 is close to a uniformly random distribution, making the separation difficult for ExCeL. Consequently, ExCeL performs better against OOD data whose predicted class rankings are sparse and random. In general this happens more in far-OOD than near-OOD, as indicated by our results.

As can be seen from Table~\ref{tab:cifar100} and Table~\ref{tab:tin}, the difference in performance between ExCeL and MaxLogit is more significant in far-OOD detection. More precisely, compared to MaxLogit, ExCeL shows 4.5\% and 5.6\% reduction in mean FPR95 for far-OOD detection in CIFAR100 and ImageNet-200, respectively. For near-OOD detection, the corresponding improvements are lower than that. They are 0.3\% and 1.9\%, respectively. This can be attributed to the relatively high semantic similarity between ID and near-OOD samples. Consequently, the class rankings in near-OOD samples tend to be more aligned with ID class rank signatures compared to far-OOD instances that are more sparse and random, rendering ExCeL more informative for differentiating between ID and far-OOD.

\section{Conclusion}
\label{sec:conclusion}

In this paper, we proposed a novel OOD score, ExCeL that combines extreme information and collective information within the output layer for enhanced OOD detection. We utilised the MaxLogit as extreme information and proposed a novel class rank score that captures information embedded across all ID classes and training samples. We demonstrated that each ID class has a unique signature, determined by the predicted classes in the subsequent ranks, which becomes less pronounced in OOD data. Experiments conducted on CIFAR100 and ImageNet-200 showed that ExCeL consistently ranks among the five top-performing methods out of twenty-one existing baselines. Furthermore, ExCeL showed the equal best performance with RMDS and GEN methods on CIFAR-100 and ImageNet-200 datasets, respectively, in terms of the overall mean rank. With regard to the overall consistent performance across datasets, ExCeL surpasses all the other post-hoc baselines.

{
    \small
    \bibliographystyle{ieeenat_fullname}
    \bibliography{main}
}

\newpage
\appendix
\section*{Appendix}

This appendix presents the per-dataset performance for near and far-OOD detection. Table~\ref{tab:cifar100_fpr} and Table~\ref{tab:cifar100_auroc} demonstrate the FPR95 and AUROC results, respectively, when CIFAR100 serves as ID. Similarly, Table~\ref{tab:tin_fpr} and Table~\ref{tab:tin_auroc} report the FPR95 and AUROC results, respectively, when ImageNet-200 is considered as ID.  

As can be seen from Table~\ref{tab:cifar100_fpr}, in terms of FPR95, ExCeL exhibits the most consistency, ranking among the top five performing methods in \textbf{five cases} (i.e., against ImageNet-200, near-OOD, SVHN, Textures, and far-OOD), which is the highest for any method when CIFAR100 is considered as ID. Similarly, when ImageNet-200 serves as ID, ExCeL ranks among the top five methods in \textbf{all seven scenarios}, as shown in Table~\ref{tab:tin_fpr}.

\vspace{5mm}

\begin{table}[h!]
 \begin{minipage}{\textwidth}
  \centering
  \tiny
  \caption{\centering FPR95 comparison of post-hoc OOD detectors for CIFAR100 (ID). The performance rank of each method is indicated within brackets. Top five values are marked in \textbf{bold}.}
  \label{tab:cifar100_fpr}
  \begin{tabular}{l|cc|c|cccc|c|}
    \toprule
    \textbf{Post-processor} & \textbf{CIFAR-10} & \textbf{ImageNet-200} & \textbf{Near-OOD} & \textbf{MNIST} & \textbf{SVHN} & \textbf{Textures} & \textbf{Places365} & \textbf{Far-OOD} \\
    \midrule
    OpenMax & 60.17 $\pm$ 0.97 (6) & 52.99 $\pm$ 0.51 (10) & 56.58 $\pm$ 0.73 (9) & 53.82 $\pm$ 4.74 (10) & 53.20 $\pm$ 1.78 (10) & 56.12 $\pm$ 1.91 (6) & \textbf{54.85 $\pm$ 1.42} (2) & 54.50 $\pm$ 0.68 (6) \\
    MSP & \textbf{58.91 $\pm$ 0.93} (3) & 50.70 $\pm$ 0.34 (6) & \textbf{54.80 $\pm$ 0.33} (3) & 57.23 $\pm$ 4.68 (15) & 59.07 $\pm$ 2.53 (16) & 61.88 $\pm$ 1.28 (10) & 56.62 $\pm$ 0.87 (6) & 58.70 $\pm$ 1.06 (12) \\
    TempScale & \textbf{58.72 $\pm$ 0.81} (1) & \textbf{50.26 $\pm$ 0.16} (5) & \textbf{54.49 $\pm$ 0.48} (2) & 56.05 $\pm$ 4.61 (14) & 57.71 $\pm$ 2.68 (14) & 61.56 $\pm$ 1.43 (9) & \textbf{56.46 $\pm$ 0.94} (5) & 57.94 $\pm$ 1.14 (11) \\
    ODIN & 60.64 $\pm$ 0.56 (8) & 55.19 $\pm$ 0.57 (13) & 57.91 $\pm$ 0.51 (10) & \textbf{45.94 $\pm$ 3.29} (2) & 67.41 $\pm$ 3.88 (19) & 62.37 $\pm$ 2.96 (12) & 59.71 $\pm$ 0.92 (11) & 58.86 $\pm$ 0.79 (13) \\
    MDS & 88.00 $\pm$ 0.49 (20) & 79.05 $\pm$ 1.22 (19) & 83.53 $\pm$ 0.60 (19) & 71.72 $\pm$ 2.94 (19) & 67.21 $\pm$ 6.09 (18) & 70.49 $\pm$ 2.48 (16) & 79.61 $\pm$ 0.34 (18) & 72.26 $\pm$ 1.56 (21) \\
    MDSEns & 95.94 $\pm$ 0.16 (22) & 95.82 $\pm$ 0.12 (22) & 95.88 $\pm$ 0.04 (22) & \textbf{2.83 $\pm$ 0.86} (1) & 82.57 $\pm$ 2.58 (22) & 84.94 $\pm$ 0.83 (20) & 96.61 $\pm$ 0.17 (22) & 66.74 $\pm$ 1.04 (17) \\
    RMDS & 61.37 $\pm$ 0.24 (12) & \textbf{49.56 $\pm$ 0.90} (2) & \textbf{55.46 $\pm$ 0.41} (5) & 52.05 $\pm$ 6.28 (6) & 51.65 $\pm$ 3.68 (8) & \textbf{53.99 $\pm$ 1.06} (4) & \textbf{53.57 $\pm$ 0.43} (1) & \textbf{52.81 $\pm$ 0.63} (3) \\
    Gram & 92.71 $\pm$ 0.64 (21) & 91.85 $\pm$ 0.86 (21) & 92.28 $\pm$ 0.29 (21) & 53.53 $\pm$ 7.45 (9) & \textbf{20.06 $\pm$ 1.96} (1) & 89.51 $\pm$ 2.54 (21) & 94.67 $\pm$ 0.60 (21) & 64.44 $\pm$ 2.37 (16) \\
    EBO & \textbf{59.21 $\pm$ 0.75} (5) & 52.03 $\pm$ 0.50 (9) & 55.62 $\pm$ 0.61 (7) & 52.62 $\pm$ 3.83 (7) & 53.62 $\pm$ 3.14 (11) & 62.35 $\pm$ 2.06 (11) & 57.75 $\pm$ 0.86 (8) & 56.59 $\pm$ 1.38 (8) \\
    OpenGAN & 78.83 $\pm$ 3.94 (16) & 74.21 $\pm$ 1.25 (17) & 76.52 $\pm$ 2.59 (16) & 63.09 $\pm$ 23.25 (17) & 70.35 $\pm$ 2.06 (21) & 74.77 $\pm$ 1.78 (18) & 73.75 $\pm$ 8.32 (16) & 70.49 $\pm$ 7.38 (19) \\
    GradNorm & 84.30 $\pm$ 0.36 (18) & 86.85 $\pm$ 0.62 (20) & 85.58 $\pm$ 0.46 (20) & 86.97 $\pm$ 1.44 (22) & 69.90 $\pm$ 7.94 (20) & 92.51 $\pm$ 0.61 (22) & 85.32 $\pm$ 0.44 (20) & 83.68 $\pm$ 1.92 (22) \\
    ReAct & 61.30 $\pm$ 0.43 (11) & 51.47 $\pm$ 0.47 (7) & 56.39 $\pm$ 0.34 (8) & 56.04 $\pm$ 5.66 (13) & 50.41 $\pm$ 2.02 (7) & \textbf{55.04 $\pm$ 0.82} (5) & \textbf{55.30 $\pm$ 0.41} (3) & \textbf{54.20 $\pm$ 1.56} (5) \\
    KLM & 84.77 $\pm$ 2.95 (19) & 71.07 $\pm$ 0.59 (16) & 77.92 $\pm$ 1.31 (17) & 73.09 $\pm$ 6.67 (20) & 50.30 $\pm$ 7.04 (6) & 81.80 $\pm$ 5.80 (19) & 81.40 $\pm$ 1.58 (19) & 71.65 $\pm$ 2.01 (20) \\
    VIM & 70.59 $\pm$ 0.43 (14) & 54.66 $\pm$ 0.42 (11) & 62.63 $\pm$ 0.27 (14) & \textbf{48.32 $\pm$ 1.07} (3) & \textbf{46.22 $\pm$ 5.46} (4) & \textbf{46.86 $\pm$ 2.29} (1) & 61.57 $\pm$ 0.77 (13) & \textbf{50.74 $\pm$ 1.00} (1) \\
    KNN & 72.80 $\pm$ 0.44 (15) & \textbf{49.65 $\pm$ 0.37} (3) & 61.22 $\pm$ 0.14 (13) & \textbf{48.58 $\pm$ 4.67} (4) & 51.75 $\pm$ 3.12 (9) & \textbf{53.56 $\pm$ 2.32} (3) & 60.70 $\pm$ 1.03 (12) & \textbf{53.65 $\pm$ 0.28} (4) \\
    DICE & 60.98 $\pm$ 1.10 (9) & 54.93 $\pm$ 0.53 (12) & 57.95 $\pm$ 0.53 (11) & \textbf{51.79 $\pm$ 3.67} (5) & \textbf{49.58 $\pm$ 3.32} (5) & 64.23 $\pm$ 1.65 (14) & 59.39 $\pm$ 1.25 (10) & 56.25 $\pm$ 0.60 (7) \\
    RankFeat & 82.78 $\pm$ 1.56 (17) & 78.40 $\pm$ 0.95 (18) & 80.59 $\pm$ 1.10 (18) & 75.01 $\pm$ 5.83 (21) & 58.49 $\pm$ 2.30 (15) & 66.87 $\pm$ 3.80 (15) & 77.42 $\pm$ 1.96 (17) & 69.45 $\pm$ 1.01 (18) \\
    ASH & 68.06 $\pm$ 0.44 (13) & 63.35 $\pm$ 0.90 (15) & 65.71 $\pm$ 0.24 (15) & 66.58 $\pm$ 3.88 (18) & \textbf{46.00 $\pm$ 2.67} (3) & 61.27 $\pm$ 2.74 (8) & 62.95 $\pm$ 0.99 (14) & 59.20 $\pm$ 2.46 (14)\\
    SHE & 60.41 $\pm$ 0.51 (7) & 57.74 $\pm$ 0.73 (14) & 59.07 $\pm$ 0.25 (12) & 58.78 $\pm$ 2.70 (16) & 59.15 $\pm$ 7.61 (17) & 73.29 $\pm$ 3.22 (17)& 65.24 $\pm$ 0.98 (15) & 64.12 $\pm$ 2.70 (15) \\
    GEN & \textbf{58.87 $\pm$ 0.69} (2) & \textbf{49.98 $\pm$ 0.05} (4) & \textbf{54.42 $\pm$ 0.33} (1) & 53.92 $\pm$ 5.71 (11) & 55.45 $\pm$ 2.76 (13) & 61.23 $\pm$ 1.40 (7) & \textbf{56.25 $\pm$ 1.01} (4) & 56.71 $\pm$ 1.59 (9) \\ 
    MaxLogit & \textbf{59.11 $\pm$ 0.64} (4) & 51.83 $\pm$ 0.70 (8) & 55.47 $\pm$ 0.66 (6) & 52.95 $\pm$ 3.82 (8) & 53.90 $\pm$ 3.04 (12) & 62.39 $\pm$ 2.13 (13) & 57.68 $\pm$ 0.91 (7) & 56.73 $\pm$ 1.33 (10) \\
    \hline
    \textbf{ExCeL (Ours)} & 61.07 $\pm$ 0.81 (10) & \textbf{49.35 $\pm$ 0.31} (1) & \textbf{55.21 $\pm$ 0.56} (4) & 54.67 $\pm$ 5.86 (12) & \textbf{45.13 $\pm$ 0.33} (2) & \textbf{51.14 $\pm$ 0.14} (2) & 58.02 $\pm$ 1.28 (9) & \textbf{52.24 $\pm$ 1.90} (2) \\
    \bottomrule
  \end{tabular}
  \end{minipage}
\end{table}

\begin{table}[h!]
 \begin{minipage}{\textwidth}
  \centering
  \tiny
  \caption{\centering AUROC comparison of post-hoc OOD detectors for CIFAR100 (ID). The performance rank of each method is indicated within brackets. Top five values are marked in \textbf{bold}.}
  \label{tab:cifar100_auroc}
  \begin{tabular}{l|cc|c|cccc|c|}
    \toprule
    \textbf{Post-processor} & \textbf{CIFAR-10} & \textbf{ImageNet-200} & \textbf{Near-OOD} & \textbf{MNIST} & \textbf{SVHN} & \textbf{Textures} & \textbf{Places365} & \textbf{Far-OOD} \\
    \midrule
    OpenMax & 74.38 $\pm$ 0.37 (14) & 78.44 $\pm$ 0.14 (15) & 76.41 $\pm$ 0.25 (15) & 76.01 $\pm$ 1.39 (17) & 82.07 $\pm$ 1.53 (9) & 80.56 $\pm$ 0.09 (6) & 79.29 $\pm$ 0.40 (10) & 79.48 $\pm$ 0.41 (11) \\
    MSP & 78.47 $\pm$ 0.07 (6) & 82.07 $\pm$ 0.17 (9) & 80.27 $\pm$ 0.11 (7) & 76.08 $\pm$ 1.86 (16) & 78.42 $\pm$ 0.89 (16) & 77.32 $\pm$ 0.71 (14) & 79.22 $\pm$ 0.29 (11) & 77.76 $\pm$ 0.44 (14) \\
    TempScale & \textbf{79.02 $\pm$ 0.06} (4) & 82.79 $\pm$ 0.09 (6) & \textbf{80.90 $\pm$ 0.07} (4) & 77.27 $\pm$ 1.85 (13) & 79.79 $\pm$ 1.05 (14) & 78.11 $\pm$ 0.72 (12) & \textbf{79.80 $\pm$ 0.25} (5) & 78.74 $\pm$ 0.51 (13) \\
    ODIN & 78.18 $\pm$ 0.14 (7) & 81.63 $\pm$ 0.08 (10) & 79.90 $\pm$ 0.11 (10) & \textbf{83.79 $\pm$ 1.31} (2) & 74.54 $\pm$ 0.76 (18) & 79.33 $\pm$ 1.08 (8) & 79.45 $\pm$ 0.26 (8) & 79.28 $\pm$ 0.21 (12)\\
    MDS & 55.87 $\pm$ 0.22 (20) & 61.50 $\pm$ 0.28 (20) & 58.69 $\pm$ 0.09 (20) & 67.47 $\pm$ 0.81 (20) & 70.68 $\pm$ 6.40 (20) & 76.26 $\pm$ 0.69 (15) & 63.15 $\pm$ 0.49 (20) & 69.39 $\pm$ 1.39 (18)\\
    MDSEns & 43.85 $\pm$ 0.31 (22) & 48.78 $\pm$ 0.19 (22) & 46.31 $\pm$ 0.24 (22) & \textbf{98.21 $\pm$ 0.78} (1) & 53.76 $\pm$ 1.63 (22) & 69.75 $\pm$ 1.14 (19) & 42.27 $\pm$ 0.73 (22) & 66.00 $\pm$ 0.69 (22) \\
    RMDS & 77.75 $\pm$ 0.19 (11) & 82.55 $\pm$ 0.02 (8) & 80.15 $\pm$ 0.11 (9) & 79.74 $\pm$ 2.49 (7) & \textbf{84.89 $\pm$ 1.10} (4) & \textbf{83.65 $\pm$ 0.51} (3) & \textbf{83.40 $\pm$ 0.46} (1) & \textbf{82.92 $\pm$ 0.42} (1)\\
    Gram & 49.41 $\pm$ 0.58 (21) & 53.91 $\pm$ 1.58 (21) & 51.66 $\pm$ 0.77 (21) & \textbf{80.71 $\pm$ 4.15} (5) & \textbf{95.55 $\pm$ 0.60} (1) & 70.79 $\pm$ 1.32 (18) & 46.38 $\pm$ 1.21 (21) & 73.36 $\pm$ 1.08 (17) \\
    EBO & \textbf{79.05 $\pm$ 0.11} (3) & 82.76 $\pm$ 0.08 (7) & \textbf{80.91 $\pm$ 0.08} (3) & 79.18 $\pm$ 1.37 (8) & 82.03 $\pm$ 1.74 (10) & 78.35 $\pm$ 0.83 (11) & 79.52 $\pm$ 0.23 (7) & 79.77 $\pm$ 0.61 (8) \\
    OpenGAN & 63.23 $\pm$ 2.44 (18) & 68.74 $\pm$ 2.29 (18) & 65.98 $\pm$ 1.26 (18) & 68.14 $\pm$ 18.78 (19) & 68.40 $\pm$ 2.15 (21) & 65.84 $\pm$ 3.43 (21) & 69.13 $\pm$ 7.08 (18) & 67.88 $\pm$ 7.16 (20) \\
    GradNorm & 70.32 $\pm$ 0.20 (17) & 69.95 $\pm$ 0.79 (17) & 70.13 $\pm$ 0.47 (17) & 65.35 $\pm$ 1.12 (21) & 76.95 $\pm$ 4.73 (17) & 64.58 $\pm$ 0.13 (22) & 69.69 $\pm$ 0.17 (17) & 69.14 $\pm$ 1.05 (19) \\
    ReAct & \textbf{78.65 $\pm$ 0.05} (5) & \textbf{82.88 $\pm$ 0.08} (5) & \textbf{80.77 $\pm$ 0.05} (5) & 78.37 $\pm$ 1.59 (11) & 83.01 $\pm$ 0.97 (8) & 80.15 $\pm$ 0.46 (7) & \textbf{80.03 $\pm$ 0.11} (3) & 80.39 $\pm$ 0.49 (6) \\
    KLM & 73.91 $\pm$ 0.25 (15) & 79.22 $\pm$ 0.28 (14) & 76.56 $\pm$ 0.25 (14) & 74.15 $\pm$ 2.59 (18) & 79.34 $\pm$ 0.44 (15) & 75.77 $\pm$ 0.45 (16) & 75.70 $\pm$ 0.24 (16) & 76.24 $\pm$ 0.52 (16)\\
    VIM & 72.21 $\pm$ 0.41 (16) & 77.76 $\pm$ 0.16 (16) & 74.98 $\pm$ 0.13 (16) & \textbf{81.89 $\pm$ 1.02} (4) & 83.14 $\pm$ 3.71 (7) & \textbf{85.91 $\pm$ 0.78} (1) & 75.85 $\pm$ 0.37 (15) & \textbf{81.70 $\pm$ 0.62} (4) \\
    KNN & 77.02 $\pm$ 0.25 (12) & \textbf{83.34 $\pm$ 0.16} (1) & 80.18 $\pm$ 0.15 (8) & \textbf{82.36 $\pm$ 1.52} (3) & 84.15 $\pm$ 1.09 (6) & \textbf{83.66 $\pm$ 0.83} (2) & 79.43 $\pm$ 0.47 (9) & \textbf{82.40 $\pm$ 0.17} (2) \\
    DICE & 78.04 $\pm$ 0.32 (10) & 80.72 $\pm$ 0.30 (11) & 79.38 $\pm$ 0.23 (11) & 79.86 $\pm$ 1.89 (6) & \textbf{84.22 $\pm$ 2.00} (5) & 77.63 $\pm$ 0.34 (13) & 78.33 $\pm$ 0.66 (13) & 80.01 $\pm$ 0.18 (7) \\
    RankFeat & 58.04 $\pm$ 2.36 (19) & 65.72 $\pm$ 0.22 (19) & 61.88 $\pm$ 1.28 (19) & 63.03 $\pm$ 3.86 (22) & 72.14 $\pm$ 1.39 (19) & 69.40 $\pm$ 3.08 (20) & 63.82 $\pm$ 1.83 (19) & 67.10 $\pm$ 1.42 (21) \\
    ASH & 76.48 $\pm$ 0.30 (13) & 79.92 $\pm$ 0.20 (12) & 78.20 $\pm$ 0.15 (13) & 77.23 $\pm$ 0.46 (14) & \textbf{85.60 $\pm$ 1.40} (3) & \textbf{80.72 $\pm$ 0.70} (5) & 78.76 $\pm$ 0.16 (12) & \textbf{80.58 $\pm$ 0.66} (5) \\
    SHE & 78.15 $\pm$ 0.03 (8) & 79.74 $\pm$ 0.36 (13) & 78.95 $\pm$ 0.18 (12) & 76.76 $\pm$ 1.07 (15) & 80.97 $\pm$ 3.98 (13) & 73.64 $\pm$ 1.28 (17) & 76.30 $\pm$ 0.51 (14) & 76.92 $\pm$ 1.16 (15) \\
    GEN & \textbf{79.38 $\pm$ 0.04} (1) & \textbf{83.25 $\pm$ 0.13} (3) & \textbf{81.31 $\pm$ 0.08} (1) & 78.29 $\pm$ 2.05 (12) & 81.41 $\pm$ 1.50 (12) & 78.74 $\pm$ 0.81 (9) & \textbf{80.28 $\pm$ 0.27} (2) & 79.68 $\pm$ 0.75 (9) \\
    MaxLogit & \textbf{79.21 $\pm$ 0.10} (2) & \textbf{82.90 $\pm$ 0.05} (4) & \textbf{81.05 $\pm$ 0.07} (2) & 78.91 $\pm$ 1.47 (10) & 81.65 $\pm$ 1.49 (11) & 78.39 $\pm$ 0.84 (10) & 79.75 $\pm$ 0.24 (6) & 79.67 $\pm$ 0.57 (10)\\
    \hline
    \textbf{ExCeL (Ours)} & 78.14 $\pm$ 0.09 (9) & \textbf{83.26 $\pm$ 0.03} (2) & 80.70 $\pm$ 0.06 (6) & 78.99 $\pm$ 1.73 (9) & \textbf{85.91 $\pm$ 0.73} (2) & \textbf{83.28 $\pm$ 0.58} (4) & \textbf{79.98 $\pm$ 0.57} (4) & \textbf{82.04 $\pm$ 0.90} (3) \\
    \bottomrule
  \end{tabular}
  \end{minipage}
\end{table}

\vfill\eject
\vspace*{2.5mm}

With regard to AUROC, when CIFAR100 is considered as ID, ExCeL ranks among the top five methods in \textbf{five cases} (i.e., against ImageNet-200, SVHN, Textures, Places365, and Far-OOD), being the most consistent method out of twenty-one baselines, as shown in Table~\ref{tab:cifar100_auroc}. When ImageNet-200 serves as ID, ExCeL drops slightly short, ranking among the top five methods only in three out of the seven scenarios according to Table~\ref{tab:tin_auroc}. Finally, ExCeL outperforms all the other baselines in terms of both FPR95 and AUROC, achieving the highest consistency (i.e., the most number of values in bold in a table) in three out of four results tables, when overall consistency is considered. This is further validated in Table~\ref{tab:cifar100} and Table~\ref{tab:tin} ({\bf cf.} Section~\ref{sec:results} in the main text) since ExCeL exhibits the best performance across both datasets in terms of the \textit{mean overall rank}.

\clearpage

\begin{table}
 \begin{minipage}{\textwidth}
  \centering
  \tiny
  \caption{\centering FPR95 comparison of post-hoc OOD detectors for ImageNet-200 (ID). The performance rank of each method is indicated within brackets. Top five values are marked in \textbf{bold}.}
  \label{tab:tin_fpr}
  \begin{tabular}{l|cc|c|ccc|c|}
    \toprule
    \textbf{Post-processor} & \textbf{SSB-hard} & \textbf{NINCO} & \textbf{Near-OOD} & \textbf{iNaturalist} & \textbf{Textures} & \textbf{OpenImage-O} & \textbf{Far-OOD} \\
    \midrule
    OpenMax & 72.37 $\pm$ 0.11 (12) & 54.59 $\pm$ 0.54 (12) & 63.48 $\pm$ 0.25 (12) & 24.53 $\pm$ 0.96 (8) & 36.80 $\pm$ 0.55 (6) & 38.03 $\pm$ 0.49 (13) & 33.12 $\pm$ 0.66 (8) \\
    MSP & \textbf{66.00 $\pm$ 0.10} (2) & \textbf{43.65 $\pm$ 0.75} (4) & \textbf{54.82 $\pm$ 0.35} (2) & 26.48 $\pm$ 0.73 (12) & 44.58 $\pm$ 0.68 (14) & 35.23 $\pm$ 0.18 (9) & 35.43 $\pm$ 0.38 (13) \\
    TempScale & \textbf{66.43 $\pm$ 0.26} (3) & \textbf{43.21 $\pm$ 0.70} (2) & \textbf{54.82 $\pm$ 0.23} (2) & 24.39 $\pm$ 0.79 (6) & 43.57 $\pm$ 0.77 (13) & 34.04 $\pm$ 0.31 (7) & 34.00 $\pm$ 0.37 (9)\\
    ODIN & 73.51 $\pm$ 0.38 (14) & 60.00 $\pm$ 0.80 (14) & 66.76 $\pm$ 0.26 (14) & \textbf{22.39 $\pm$ 1.87} (3) & 42.99 $\pm$ 1.56 (12) & 37.30 $\pm$ 0.59 (12) & 34.23 $\pm$ 1.05 (11) \\
    MDS & 83.65 $\pm$ 0.47 (18) & 74.57 $\pm$ 0.15 (17) & 79.11 $\pm$ 0.31 (17) & 58.53 $\pm$ 0.75 (17) & 58.16 $\pm$ 0.84 (17) & 68.29 $\pm$ 0.28 (18) & 61.66 $\pm$ 0.27 (17) \\
    MDSEns & 92.13 $\pm$ 0.05 (22) & 91.36 $\pm$ 0.16 (21) & 91.75 $\pm$ 0.10 (21) & 83.37 $\pm$ 0.70 (20) & 72.27 $\pm$ 0.48 (20) & 87.26 $\pm$ 0.10 (21) & 80.96 $\pm$ 0.38 (20) \\
    RMDS & \textbf{65.91 $\pm$ 0.27} (1) & \textbf{42.13 $\pm$ 1.04} (1) & \textbf{54.02 $\pm$ 0.58} (1) & 24.70 $\pm$ 0.90 (9) & 37.80 $\pm$ 1.32 (7) & 34.85 $\pm$ 0.31 (8) & 32.45 $\pm$ 0.79 (7) \\
    Gram & 85.68 $\pm$ 0.85 (19) & 87.13 $\pm$ 1.89 (20) & 86.40 $\pm$ 1.21 (20) & 85.54 $\pm$ 0.40 (21) & 80.87 $\pm$ 1.20 (21) & 86.66 $\pm$ 1.27 (20) & 84.36 $\pm$ 0.78 (21) \\
    EBO & 69.77 $\pm$ 0.32 (7) & 50.70 $\pm$ 0.89 (9) & 60.24 $\pm$ 0.57 (9) & 26.41 $\pm$ 2.29 (11) & 41.43 $\pm$ 1.85 (10) & 36.74 $\pm$ 1.14 (11) & 34.86 $\pm$ 1.30 (12) \\
    OpenGAN & 88.07 $\pm$ 2.23 (20) & 80.23 $\pm$ 5.71 (18) & 84.15 $\pm$ 3.85 (19) & 60.13 $\pm$ 9.79 (18) & 66.00 $\pm$ 9.97 (18) & 66.34 $\pm$ 8.44 (17) & 64.16 $\pm$ 9.33 (18) \\
    GradNorm & 82.17 $\pm$ 0.62 (17) & 83.17 $\pm$ 0.21 (19) & 82.67 $\pm$ 0.30 (18) & 61.31 $\pm$ 2.86 (19) & 66.88 $\pm$ 3.59 (19) & 71.16 $\pm$ 0.23 (19) & 66.45 $\pm$ 0.22 (19) \\
    ReAct & 71.51 $\pm$ 1.92 (10) & 53.47 $\pm$ 2.46 (11) & 62.49 $\pm$ 2.19 (11) & \textbf{22.97 $\pm$ 2.25} (5) & \textbf{29.67 $\pm$ 1.35} (4) & \textbf{32.86 $\pm$ 0.74} (2) & \textbf{28.50 $\pm$ 0.95} (5) \\
    KLM & 78.19 $\pm$ 2.30 (16) & 62.33 $\pm$ 2.66 (16) & 70.26 $\pm$ 0.64 (16) & 26.66 $\pm$ 1.61 (13) & 50.24 $\pm$ 1.26 (16) & 45.81 $\pm$ 0.59 (15) & 40.90 $\pm$ 1.08 (15) \\
    VIM & 71.28 $\pm$ 0.49 (9) & 47.10 $\pm$ 1.10 (7) & 59.19 $\pm$ 0.71 (6) & 27.34 $\pm$ 0.38 (14) & \textbf{20.39 $\pm$ 0.17} (1) & 33.86 $\pm$ 0.63 (6) & \textbf{27.20 $\pm$ 0.30} (1) \\
    KNN & 73.71 $\pm$ 0.31 (15) & 46.64 $\pm$ 0.73 (6) & 60.18 $\pm$ 0.52 (8) & 24.46 $\pm$ 1.06 (7) & \textbf{24.45 $\pm$ 0.29} (2) & \textbf{32.90 $\pm$ 1.12} (3) & \textbf{27.27 $\pm$ 0.75} (2) \\
    DICE & 70.84 $\pm$ 0.30 (8) & 52.91 $\pm$ 1.20 (10) & 61.88 $\pm$ 0.67 (10) & 29.66 $\pm$ 2.62 (15) & 40.96 $\pm$ 1.87 (8) & 38.91 $\pm$ 1.16 (14) & 36.51 $\pm$ 1.18 (14) \\
    RankFeat & 90.79 $\pm$ 0.37 (21) & 93.32 $\pm$ 0.11 (22) & 92.06 $\pm$ 0.23 (22) & 98.00 $\pm$ 0.80 (22) & 99.40 $\pm$ 0.68 (22) & 95.77 $\pm$ 0.85 (22) & 97.72 $\pm$ 0.75 (22) \\
    ASH & 72.14 $\pm$ 0.97 (11) & 57.63 $\pm$ 0.98 (13) & 64.89 $\pm$ 0.90 (13) & \textbf{22.49 $\pm$ 2.24} (4) & \textbf{25.65 $\pm$ 0.80} (3) & \textbf{33.72 $\pm$ 0.97} (5) & \textbf{27.29 $\pm$ 1.12} (3) \\
    SHE & 72.64 $\pm$ 0.30 (13) & 60.96 $\pm$ 1.33 (15) & 66.80 $\pm$ 0.74 (15) & 34.38 $\pm$ 3.48 (16) & 45.58 $\pm$ 2.42 (15) & 46.54 $\pm$ 1.34 (16) & 42.17 $\pm$ 1.24 (16) \\
    GEN & \textbf{66.79 $\pm$ 0.26} (4) & \textbf{43.61 $\pm$ 0.61} (3) & \textbf{55.20 $\pm$ 0.20} (4) & \textbf{22.03 $\pm$ 0.98} (1) & 42.01 $\pm$ 0.92 (11) & \textbf{32.25 $\pm$ 0.31} (1) & 32.10 $\pm$ 0.59 (6) \\
    MaxLogit & 69.64 $\pm$ 0.37 (6) & 49.87 $\pm$ 0.94 (8) & 59.76 $\pm$ 0.59 (7) & 25.09 $\pm$ 2.04 (10) & 41.25 $\pm$ 1.86 (9) & 35.76 $\pm$ 0.74 (10) & 34.03 $\pm$ 1.21 (10) \\
    \hline
    \textbf{ExCeL (Ours)} & \textbf{69.28 $\pm$ 0.60} (5) & \textbf{46.51 $\pm$ 0.20} (5) & \textbf{57.90 $\pm$ 0.40} (5) & \textbf{22.29 $\pm$ 1.00} (2) & \textbf{30.14 $\pm$ 0.64} (5) & \textbf{32.91 $\pm$ 0.76} (4) & \textbf{28.45 $\pm$ 0.80} (4) \\
    \bottomrule
  \end{tabular}
  \end{minipage}
\end{table}

\begin{table}
 \begin{minipage}{\textwidth}
  \centering
  \tiny
  \caption{\centering AUROC comparison of post-hoc OOD detectors for ImageNet-200 (ID). The performance rank of each method is indicated within brackets. Top five values are marked in \textbf{bold}.}
  \label{tab:tin_auroc}
  \begin{tabular}{l|cc|c|ccc|c|}
    \toprule
    \textbf{Post-processor} & \textbf{SSB-hard} & \textbf{NINCO} & \textbf{Near-OOD} & \textbf{iNaturalist} & \textbf{Textures} & \textbf{OpenImage-O} & \textbf{Far-OOD} \\
    \midrule
    OpenMax & 77.53 $\pm$ 0.08 (13) & 83.01 $\pm$ 0.17 (15) & 80.27 $\pm$ 0.10 (13) & 92.32 $\pm$ 0.32 (11) & 90.21 $\pm$ 0.07 (12) & 88.07 $\pm$ 0.14 (13) & 90.20 $\pm$ 0.17 (12) \\
    MSP & \textbf{80.38 $\pm$ 0.03} (3) & \textbf{86.29 $\pm$ 0.11} (3) & \textbf{83.34 $\pm$ 0.06} (3) & 92.80 $\pm$ 0.25 (9) & 88.36 $\pm$ 0.13 (14) & 89.24 $\pm$ 0.02 (9) & 90.13 $\pm$ 0.09 (13) \\
    TempScale & \textbf{80.71 $\pm$ 0.02} (2) & \textbf{86.67 $\pm$ 0.08} (1) & \textbf{83.69 $\pm$ 0.04} (1) & 93.39 $\pm$ 0.25 (7) & 89.24 $\pm$ 0.11 (13) & 89.84 $\pm$ 0.02 (6) & 90.82 $\pm$ 0.09 (10) \\
    ODIN & 77.19 $\pm$ 0.06 (14) & 83.34 $\pm$ 0.12 (13) & 80.27 $\pm$ 0.08 (13) & \textbf{94.37 $\pm$ 0.41} (2) & 90.65 $\pm$ 0.20 (8) & \textbf{90.11 $\pm$ 0.15} (5) & \textbf{91.71 $\pm$ 0.19} (5) \\
    MDS & 58.38 $\pm$ 0.58 (20) & 65.48 $\pm$ 0.46 (19) & 61.93 $\pm$ 0.51 (19) & 75.03 $\pm$ 0.76 (19) & 79.25 $\pm$ 0.33 (20) & 69.87 $\pm$ 0.14 (19) & 74.72 $\pm$ 0.26 (18) \\
    MDSEns & 50.46 $\pm$ 0.36 (22) & 58.18 $\pm$ 0.42 (21) & 54.32 $\pm$ 0.24 (22) & 62.16 $\pm$ 0.73 (21) & 80.70 $\pm$ 0.48 (18) & 64.96 $\pm$ 0.51 (21) & 69.27 $\pm$ 0.57 (21) \\
    RMDS & \textbf{80.20 $\pm$ 0.23} (4) & 84.94 $\pm$ 0.28 (9) & \textbf{82.57 $\pm$ 0.25} (5) & 90.64 $\pm$ 0.46 (16) & 86.77 $\pm$ 0.38 (15) & 86.77 $\pm$ 0.22 (16) & 88.06 $\pm$ 0.34 (16) \\
    Gram & 65.95 $\pm$ 1.08 (18) & 69.40 $\pm$ 1.07 (18) & 67.67 $\pm$ 1.07 (18) & 65.30 $\pm$ 0.20 (20) & 80.53 $\pm$ 0.37 (19) & 67.72 $\pm$ 0.58 (20) & 71.19 $\pm$ 0.24 (20) \\
    EBO & 79.83 $\pm$ 0.02 (6) & 85.17 $\pm$ 0.11 (8) & 82.50 $\pm$ 0.05 (6) & 92.55 $\pm$ 0.50 (10) & 90.79 $\pm$ 0.16 (7) & 89.23 $\pm$ 0.26 (10) & 90.86 $\pm$ 0.21 (9) \\
    OpenGAN & 55.08 $\pm$ 1.84 (21) & 64.49 $\pm$ 4.98 (20) & 59.79 $\pm$ 3.39 (20) & 75.32 $\pm$ 3.32 (18) & 70.58 $\pm$ 4.66 (21) & 73.54 $\pm$ 4.48 (18) & 73.15 $\pm$ 4.07 (19) \\
    GradNorm & 72.12 $\pm$ 0.43 (17) & 73.39 $\pm$ 0.63 (17) & 72.75 $\pm$ 0.48 (17) & 86.06 $\pm$ 1.90 (17) & 86.07 $\pm$ 0.36 (17) & 80.66 $\pm$ 1.09 (17) & 84.26 $\pm$ 0.87 (17) \\
    ReAct & 78.97 $\pm$ 1.33 (10) & 84.76 $\pm$ 0.64 (10) & 81.87 $\pm$ 0.98 (9) & 93.65 $\pm$ 0.88 (6) & \textbf{92.86 $\pm$ 0.47} (4) & \textbf{90.40 $\pm$ 0.35} (2) & \textbf{92.31 $\pm$ 0.56} (3) \\
    KLM & 77.56 $\pm$ 0.18 (12) & 83.96 $\pm$ 0.12 (12) & 80.76 $\pm$ 0.08 (12) & 91.80 $\pm$ 0.21 (13) & 86.13 $\pm$ 0.12 (16) & 87.66 $\pm$ 0.17 (14) & 88.53 $\pm$ 0.11 (15) \\
    VIM & 74.04 $\pm$ 0.31 (16) & 83.32 $\pm$ 0.19 (14) & 78.68 $\pm$ 0.24 (16) & 90.96 $\pm$ 0.36 (15) & \textbf{94.61 $\pm$ 0.12} (3) & 88.20 $\pm$ 0.18 (12) & 91.26 $\pm$ 0.19 (7) \\
    KNN & 77.03 $\pm$ 0.23 (15) & \textbf{86.10 $\pm$ 0.12} (4) & 81.57 $\pm$ 0.17 (11) & \textbf{93.99 $\pm$ 0.36} (3) & \textbf{95.29 $\pm$ 0.02} (1) & \textbf{90.19 $\pm$ 0.32} (3) & \textbf{93.16 $\pm$ 0.22} (2) \\
    DICE & 79.06 $\pm$ 0.05 (9) & 84.49 $\pm$ 0.24 (11) & 81.78 $\pm$ 0.14 (10) & 91.81 $\pm$ 0.79 (12) & 91.53 $\pm$ 0.21 (6) & 89.06 $\pm$ 0.34 (11) & 90.80 $\pm$ 0.31 (11) \\
    RankFeat & 58.74 $\pm$ 0.94 (19) & 55.10 $\pm$ 2.52 (22) & 56.92 $\pm$ 1.59 (21) & 33.08 $\pm$ 4.68 (22) & 29.10 $\pm$ 2.57 (22) & 52.48 $\pm$ 4.44 (22) & 38.22 $\pm$ 3.85 (22) \\
    ASH & 79.52 $\pm$ 0.37 (7) & 85.24 $\pm$ 0.08 (7) & 82.38 $\pm$ 0.19 (8) & \textbf{95.10 $\pm$ 0.47} (1) & \textbf{94.77 $\pm$ 0.19} (2) & \textbf{91.82 $\pm$ 0.25} (1) & \textbf{93.90 $\pm$ 0.27} (1) \\
    SHE & 78.30 $\pm$ 0.20 (11) & 82.07 $\pm$ 0.33 (16) & 80.18 $\pm$ 0.25 (15) & 91.43 $\pm$ 1.28 (14) & 90.51 $\pm$ 0.19 (10) & 87.49 $\pm$ 0.70 (15) & 89.81 $\pm$ 0.61 (14) \\
    GEN & \textbf{80.75 $\pm$ 0.03} (1) & \textbf{86.60 $\pm$ 0.08} (2) & \textbf{83.68 $\pm$ 0.06} (2) & \textbf{93.70 $\pm$ 0.18} (5) & 90.25 $\pm$ 0.10 (11) & \textbf{90.13 $\pm$ 0.06} (4) & 91.36 $\pm$ 0.10 (6) \\
    MaxLogit & \textbf{80.15 $\pm$ 0.01} (5) & \textbf{85.65 $\pm$ 0.09} (5) & \textbf{82.90 $\pm$ 0.04} (4) & 93.12 $\pm$ 0.45 (8) & 90.60 $\pm$ 0.16 (9) & 89.62 $\pm$ 0.21 (8) & 91.11 $\pm$ 0.19 (8) \\
    \hline
    \textbf{ExCeL (Ours)} & 79.39 $\pm$ 0.03 (8) & 85.40 $\pm$ 0.04 (6) & 82.40 $\pm$ 0.04 (7) & \textbf{93.76 $\pm$ 0.43} (4) & \textbf{92.40 $\pm$ 0.05} (5) & 89.75 $\pm$ 0.32 (7) & \textbf{91.97 $\pm$ 0.27} (4) \\
    \bottomrule
  \end{tabular}
  \end{minipage}
\end{table}

\end{document}